\documentclass[journal]{IEEEtran}
\usepackage{amsmath,amssymb,amsthm,amscd,verbatim,graphicx}
\usepackage[mathscr]{eucal}
\usepackage{bbm} 
\usepackage{enumerate} 
\usepackage{enumitem} 
\usepackage{floatrow} 
\usepackage{tikz}
\usepackage{float} 
\usepackage{cprotect} 
\usepackage{caption}
\usepackage[framemethod=tikz]{mdframed} 
\usepackage{xparse, adforn} 
\usepackage{listings} 
\usepackage{color} 
\usepackage{bm} 
\usepackage{xcolor} 
\usepackage{hyperref}
\hypersetup{ 
    colorlinks=true, 
    linkcolor=blue,
    filecolor=magenta,      
    urlcolor=blue,
    citecolor=blue
}
\usepackage{cleveref}
\usepackage[nolist,nohyperlinks]{acronym}
\usepackage{subcaption}

\acrodef{AI}{Artificial Intelligence}
\acrodef{XAI}{Explainable Artificial Intelligence}
\acrodef{ML}{Machine Learning}
\acrodef{TU}{Transferable Utility}
\acrodef{SHAP}{SHapley Additive exPlanations}
\acrodef{SAGE}{Shapley Additive Global importancE}
\acrodef{DGP}{Data Generating Process}
\acrodef{AIC}{Akaike Information Criterion}
\acrodef{BIC}{Bayesian Information Criterion}
\acrodef{OLS}{Ordinary Least Squares}

\usepackage{babel}
\providecommand{\lemmaname}{Lemma}
\providecommand{\remarkname}{Remark}
\providecommand{\theoremname}{Theorem}
\providecommand{\definitionname}{Definition}
\theoremstyle{plain}

\theoremstyle{remark}

\theoremstyle{plain}

\theoremstyle{definition}
\newtheorem{definition}{\protect\definitionname}[section]
\newtheorem{axiom}{Axiom}
\newtheorem{example}{Example}

\newcommand{\ex}[1]{\char`\\ \{#1\}}
\newcommand{\R}{\mathbb{R}}

\newcommand{\set}[1]{{\left\{#1\right\}}}

\newcommand{\deq}{\mathbin{\raise0.4pt\hbox{:}{=}}}
\newcommand{\eps}{\varepsilon}
 \newcommand{\ind}{\perp\!\!\!\!\perp}

\newcommand{\fullmodel}{F}

\newcommand{\submodel}{S}
\newcommand{\submodeltwo}{T}
\newcommand{\dimension}{d}
\newcommand{\CF}{C}
\newcommand{\CFtwo}{K}
\newcommand{\MC}{M}
\newcommand{\shapley}{\varphi}
\newcommand{\pone}{i}
\newcommand{\ptwo}{j}

\usepackage{cite}
\usepackage{algpseudocode}
\usepackage{algorithm}
\begin{document}
\title{Shapley values for feature selection:\\The good, the bad, and the axioms}
\author{Daniel~Fryer$^1$,
        Inga~Str\"umke$^2$,
        and~Hien~Nguyen$^3$
\thanks{$^1$School of Mathematics and Physics, The University of Queensland, St Lucia, Australia}
\thanks{$^2$SimulaMet, Simula Research Laboratory, Oslo, Norway}
\thanks{$^3$Department of Mathematics and Statistics, La Trobe University, Melbourne, Australia}}

\maketitle

\begin{abstract}
The Shapley value has become popular in the Explainable AI (XAI) literature, thanks, to a large extent, to a solid theoretical foundation, including four ``favourable and fair" axioms for attribution in transferable utility games. The Shapley value is provably the only solution concept satisfying these axioms. In this paper, we introduce the Shapley value and draw attention to its recent uses as a feature selection tool. We call into question this use of the Shapley value, using simple, abstract `toy' counterexamples to illustrate that the axioms may work against the goals of feature selection. From this, we develop a number of insights that are then investigated in concrete simulation settings, with a variety of Shapley value formulations, including SHapley Additive exPlanations (SHAP) and Shapley Additive Global importancE (SAGE).
\end{abstract}

\section{Introduction}
\IEEEPARstart{T}{he} problem of feature selection in \ac{ML} constitutes selecting some subset $\submodel$ of a set $\fullmodel$ of $|\fullmodel| = \dimension$ feature indices, such that the submodel formed from the features indexed by $\submodel$ will maximise some evaluation function $\CF(\submodel)$ of the submodel, while minimising a cost (or complexity), which is increasing in $|\submodel|$. The model chosen by this procedure is the \textit{selected} model. 

A similar (and more general) problem -- model selection -- has deep roots in computational statistics \cite{[G10]}, where attention is paid to inferential nuances like quantification of uncertainty, significance testing, confounding predictors, collinearity, and the design of experiments. It was in this literature that the Shapley value was first applied to linear regression models, with its own history of discourse (see \cite{[S2],[S9],[S10],[S14],[S15],fryer2020shapleyCI} and the more critical \cite{[C5]}, which traces development to \cite{[H1],[H2],[H3]}, with reinventions by \cite{[S7]} and \cite{[H4]}).

The Shapley value has, over recent years, become a popular method for interpretable feature attribution in fitted \ac{ML} models (cf. \cite{[S1],[S3],[S4],[S5],[S6],[S8],[S11],[S12],[S13],[S15],[S16],[S17],[F4],fryer2020explaining}), holding promise for the development of \ac{XAI}. Attribution (or credit allocation), here, is the determination of the contribution by each feature to the performance of a model -- often the selected model. The \ac{ML} methods that stand out in terms of popularity are \ac{SHAP} \cite{[S5],[S3]}, Shapley Effects \cite{[S15]} and \ac{SAGE} \cite{[S17]}, though the Shapley value itself carries a rich history of investigation in the context of game theory -- Lloyd Shapley's 1953 seminal paper \cite{[G9]} has over 9000 citations, and the concept has attracted the attention of various Nobel prize winning economists \cite{[G1],[G2],[G3],[G4],[G5],[G6],[G7]}. 

Particular praise is given, in both the game theory and \ac{ML} literature, to a small set of ``favourable and fair" axioms, commonly known as \textit{efficiency}, \textit{null player}, \textit{symmetry} and \textit{additivity}, under which the Shapley value can be uniquely defined. We will introduce these axioms in~\cref{sec:axioms}. While much attention has been paid by game theorists and economists to interpreting and reformulating the axioms, and towards investigating axiom sets (and game formulations) that lead to Shapley value alternatives~\cite{[G1],[G2],[G4]}, we found comparatively little attention to these matters in \ac{ML}~\cite{[A4],[C1],[S17]}. There have, however, been a number of recent criticisms of the Shapley value in \ac{ML}~\cite{[C2],[C3],[C4]}, suggesting to us that the Shapley value and its alternatives may be further developed considerably over the coming years.

This paper is not an exhaustive theoretical or empirical study of the worth of various Shapley value methods in \ac{ML}, neither in general nor for feature selection. Our goal is to draw scrutiny towards the Shapley value axioms, and attention towards the generality of the game theoretic formulation. We do this in the specific context of feature selection, since this direction is particularly underdeveloped, and because in both academic and industrial settings we have encountered what we consider to be an over-reliance on axiomatic ``guarantees" (e.g., of ``fairness") when appropriating Shapley based feature \textit{attribution} methods for feature \textit{selection} (see, e.g.,~\cite{[F2],[F3],[F5],[F6],[F7],[F8],[F9],[F10],[F14],[F15]}). Through the arguments in this paper, the authors are convinced of two things:
\begin{itemize}
    \item The axioms do not \textit{in general} provide any guarantee that the Shapley value is suited to feature selection, and may, in some cases, imply the opposite.
    \item The relevance of the Shapley value to the feature selection task (indeed, to any \ac{ML} task) is governed by the specific game formulation, and the justification of this relevance from the axioms is non-trivial.   
\end{itemize}

In \Cref{sec:background}, we introduce the Shapley value and game formulation (\Cref{sec:shapley}), the axioms (\Cref{sec:axioms}), we give a brief overview of feature selection in general (\Cref{sec:fselect}), and then a brief overview of Shapley value feature selection (\Cref{sec:sfselect}). In  \Cref{sec:mean}, we investigate the significance of the Shapley value axioms in the context of feature selection, introducing general ``toy'' examples to illustrate. Then, in~\Cref{sec:experiments} we perform simulation studies on more concrete data sets, with various evaluation functions and game formulations, including the popular mean absolute \ac{SHAP} and \ac{SAGE} formulations.
 
\begin{quote}
    ``\textit{Far better an approximate answer to the right question, which is often vague, than an exact answer to the wrong question\ldots}" -- John Tukey \cite{tukey1962future}.
\end{quote}

\section{Background knowledge} \label{sec:background}
\subsection{The Shapley value} \label{sec:shapley}
In~\Cref{def:tu}, we define a \ac{TU} game. This definition captures the general scenario where a set of objects, denoted by $\fullmodel$, has some associated evaluation, $\CF(\fullmodel)$, and, for any subset $\submodel$ of $\fullmodel$, the evaluation $\CF(\submodel)$ is also well-defined. This captures a typical feature selection scenario, in which $\fullmodel = \{1,\ldots,\dimension\}$ represents the indices of all features in the full model of dimension $\dimension$, and $\submodel$ represents the indices of features in some submodel of dimension $|\submodel|$. In a \ac{TU} game, we are guaranteed that the worth of every submodel can be evaluated, as $\CF(\submodel)$. 

In the following, we use $2^\fullmodel$ to denote the set of all possible subsets of the objects $\fullmodel$, which include $\fullmodel$ itself and the empty set, denoted by $\varnothing$.
\begin{definition}[TU game]\label{def:tu}
A \ac{TU} game is a pair $(\fullmodel,\CF)$, where $\fullmodel = \set{1,\ldots,\dimension}$ is a set of indices called \textit{players} and the \textit{characteristic function} $\CF:2^\fullmodel \rightarrow \R$ assigns a non-negative real value $\CF(\submodel)$ to every coalition $\submodel \subseteq \fullmodel$. Furthermore, $\CF$ assigns the value zero to the empty coalition $\varnothing$, i.e. $\CF(\varnothing) = 0$.
\end{definition}
For readability, we will often refer to $\CF$ as an \textit{evaluation function}, since it evaluates the worth of each coalition, and we will refer to the players as \textit{features}.

The Shapley value (\Cref{def:shapley}) is intuitively appealing for feature selection. At first glance, it extracts (or compresses) information from the evaluation function $\CF$, to assign a single value $\shapley_i$ representing the worth of each feature in the modelling task. Here, the meaning of \textit{worth} depends strongly on the choice of evaluation function, and (less obviously) on the manner in which features are understood to be removed from the model (see \cite{[C1]} and \cite[Section 4]{[S17]}). Generalising the terminology in~\cite{[C1]}, we refer to these choices as the Shapley value \textit{game formulation}. 

An attractive notion is that, with a suitable game formulation, the Shapley value could guarantee a principled method of feature selection. However, while such a method of feature selection is \textit{principled} via the axioms that are discussed in~\Cref{sec:axioms}, it is, as we will see, the principles (or axioms) themselves that are not \textit{in general} suited to feature selection, and must be scrutinised in both the context of the specific game formulation, and the context of the outcome that is desired from the feature selection task.

Central to the definition of the Shapley value is the notion of a marginal contribution, which can be understood as the amount by which the evaluation of a given submodel increases, upon introducing a given feature to the submodel.
\begin{definition}[Marginal contribution]
The marginal contribution of feature $i$ to submodel $S$ is defined as the difference in evaluation when $i$ is added to the submodel:
\[
\MC_\pone(\submodel) = \CF(\submodel \cup \set{\pone}) - \CF(\submodel).
\]
\end{definition}

\begin{definition}[The Shapley value] \label{def:shapley}
The Shapley value of feature $i$ is defined as a weighted average over all marginal contributions by feature $i$. That is, over $\MC_\pone(\submodel)$ for every subset $\submodel$ of $\fullmodel$ that excludes $\pone$.
\begin{equation} \label{eq:shapley}
\shapley_\pone = \sum_{\submodel \in 2^{\fullmodel\ex{\pone}}} \omega(\submodel) \MC_\pone(\submodel),
\end{equation}
where $\omega(\submodel) = |\submodel|!(|\fullmodel| - |\submodel| - 1)!/|\fullmodel|!$ are the specific weights that define the Shapley value.
\end{definition}

The formula for the Shapley value, or at least that for its weights, may not immediately lend itself to intuition. Historically, much attention has been paid instead to the small set of axioms (in~\Cref{sec:axioms}) from which~\Cref{def:shapley} can be \textit{uniquely} derived. An \textit{axiom} is a principle, usually taken to be self evident, from which other truths may be derived. The simple and intuitive nature of the Shapley value axioms encourages an assessment that the Shapley value is an \textit{explainable} or \textit{interpretable} approach to computing the importance of features. However, great care must be exercised in establishing the exact meaning of \textit{importance}, both in the general Shapley value context \textit{and} in the specific contexts in which the Shapley value is applied.

\subsection{The Shapley value axioms} \label{sec:axioms}
In keeping with the machine learning literature, and as a matter of preference, we present the following four axioms as a unique characterisation of the Shapley value. However, there are a number of alternative axiomatisations available that may provide varying levels of insight \cite{[G1]}. 

\begin{axiom}[\textit{Efficiency}] \label{ax:eff}
In a \ac{TU} game $(\CF, \fullmodel)$, the worth of the full model $\CF(\fullmodel)$ is distributed in a lossless manner among the features:
\[
\textstyle{ \sum_{\pone \in \fullmodel} \shapley_\pone = \CF(\fullmodel)}.
\]
\end{axiom}

\begin{axiom}[\textit{Null player}] \label{ax:nul}
In a \ac{TU} game $(\CF, \fullmodel)$, if feature $\pone$ contributes nothing to each submodel it enters, then its Shapley value is zero:
\[
[ (\forall \,\submodel )\; \CF(\submodel \cup \{\pone\}) = \CF(\{\pone\}) ] \implies \shapley_\pone = 0.
\]
\end{axiom}

\begin{axiom}[\textit{Symmetry}] \label{ax:sym}
In a \ac{TU} game $(\CF, \fullmodel)$, any two features $\pone, \ptwo$ that play equal roles have equal Shapley values:
\[
[ (\forall \,\submodel\ex{\pone,\ptwo}) \; \CF(\submodel \cup \set{\pone}) = \CF(\submodel \cup \set{\ptwo}) ] \implies \shapley_\pone = \shapley_\ptwo.
\]
\end{axiom}

\begin{axiom}[\textit{Additivity}] \label{ax:add}
Given two \ac{TU} games $(\CF,\fullmodel), (\CFtwo, \fullmodel)$, the Shapley value of feature $\pone$ preserves addition of the evaluation functions:
\[
\shapley_\pone(\CF + \CFtwo) = \shapley_\pone(C) + \shapley_\pone(\CFtwo).
\]
\end{axiom}

In~\Cref{ax:add}, the notation $\shapley_\pone(\CF)$ denotes the Shapley value of player $\pone$ using the evaluation function $\CF$, and the addition of two evaluation functions is defined as the natural (pointwise) addition $(C+K)(S) = C(S) + K(S).$

Axioms~\ref{ax:nul}--\ref{ax:add} can be replaced (as in~\cite{[G7]}) by the following single axiom, which~\cite{[G6]} named \textit{balanced contributions}.

\begin{axiom}[\textit{Balanced contributions}] \label{ax:bal}
Let $\CF_{\pone}$ denote the game produced by restricting the feature set $\fullmodel$ to $\fullmodel\ex{\pone}$. Then,
\[
    \shapley_\pone(\CF) - \shapley_\pone(\CF_{\ptwo}) = \shapley_\ptwo(\CF) - \shapley_\ptwo(\CF_{\pone}) \,.
\]
\end{axiom}

To paraphrase~\cite{[G6]}, balanced contributions is a principle of fairness in cooperation. It states that every pair of features should share equally the gain (or loss) received from their cooperation.

In a \ac{TU} game, Axioms~\ref{ax:eff}--\ref{ax:add} are sufficient to uniquely define the \textit{exact} Shapley value. However, this value is only unique up to the choice of characteristic function and game formulation. Between such choices, the Shapley value will vary greatly. Also, it should be noted that in many practical settings, the Shapley value is only approximated, since the complexity of \eqref{eq:shapley} is exponential in the number of features (see, e.g., \cite{[S3]} and \cite[Section 6.2]{[S17]}).  

In algorithmic feature selection the search space generally involves $2^d$ submodels. Existing feature selection methods all take some approach to avoiding the exhaustive search of the model space.

\subsection{Feature selection in general} \label{sec:fselect}

If an evaluation function $\CF$ is used for feature selection, it should correspond to a specific goal. Feature selection may target a number of typically exclusive goals, between which there is generally understood to be a trade-off in performance. These goals include, but may not be limited to, succinctly describing a data generating process; improving the predictive performance of the model; maximising the overall significance or power of the model, or of its parameters, with respect to a hypothesis; and producing a more cost-effective or computationally efficient predictor. There are two prominent categories of feature selection techniques identified by the surveys of~\cite{[F11],[F12]}: 

\begin{itemize}
    \item \textit{Wrapper methods} evaluate a number of trained submodels selected via sequential (e.g., stepwise forward/backward) elimination, or via a heuristic search algorithm. While these model-driven evaluations are extrinsic to the data, they are native to the specific modelling task.
    \item \textit{Filter methods} are general model-independent frameworks that avoid the computational burden of model training, present in wrapper methods. These methods rank features via empirical estimates of intrinsic properties of the data, such as covariance or mutual information.
\end{itemize}
 
Both of the above methods can be applied in the context of Shapley values, given appropriate choices for the evaluation function and game formulation (see \Cref{sec:sfselect}). Regardless of the specific goal, the feature selection task is motivated by the notion that a submodel exists that is of higher value than the full model, in the sense of \Cref{def:monotonic}.

\begin{definition}[Monotonicity] \label{def:monotonic}
In a \ac{TU} game $(\fullmodel, \CF)$, the evaluation function $\CF:2^\fullmodel \rightarrow \R$ is called monotonic when $C(S)$ does not decrease whenever new features are added to $S$.
\[
(\forall \, \submodel, \submodeltwo \subseteq \fullmodel) \;\submodel \subset \submodeltwo \implies \CF(\submodel) \leq \CF(\submodeltwo).   
\]
\end{definition}

Popular examples of non-monotonic evaluation functions in feature selection are the \ac{AIC} and \ac{BIC}, often used in conjunction with a stepwise procedure~\cite{[G10]}.

\subsection{Shapley values for feature selection} \label{sec:sfselect}

The following is the simplest general Shapley value feature selection procedure. 
\begin{algorithm}[H]
\caption{Attribution selection}\label{alg:attsel}
\begin{algorithmic}[1]
    \State Choose an objective function $\CF$.
    \State Compute the Shapley value $\shapley_\pone$ for all features $\pone \in \fullmodel$.
    \State Select the $k$ highest ranking features, for $k < d = |F|$.
\end{algorithmic}
\end{algorithm}
An alternative to the final step in \Cref{alg:attsel} is to select features $\pone$ for which $\shapley_\pone$ lies above some threshold. Upon review of the literature, we found a number of articles suggesting to use a variant of \Cref{alg:attsel} \cite{[F7],[F8],[F14],[F15]}, two completely applied uses of \Cref{alg:attsel} \cite{[F9],[F10]}, and one paper critical of the algorithm \cite{[F1]}.

Alternatives to~\Cref{alg:attsel} are found in~\cite{[F5],[F2],[F3],[F6]}. In \cite{[F2],[F3]} the considered coalition sizes are restricted, and a stepwise selection procedure is performed. A genetic algorithm is described in~\cite{[F5]}. In~\cite{[F6]}, the problem of model averaging (see~\Cref{sec:mean-avg}) is discussed, and it is claimed that the method avoids the influence of unselected features via a decomposition of the Shapley value into high-order interaction components. We do not investigate these methods, but we expect that controlling the coalition sizes may have a positive impact, at least on the problem considered in~\Cref{ex:secret} and~\Cref{sim:secret}.

\section{The meaning of the axioms} \label{sec:mean}

In this section we use ``toy" examples to interrogate some general consequences of the axioms, and to gain insight into how feature selection may be impacted by them. The insights discussed in this section are too general for drawing conclusions about the value of any specific Shapley value feature selection procedure, in practice.

\subsection{The meaning of model averaging} \label{sec:mean-avg}

As a consequence of Axioms~\ref{ax:eff} and~\ref{ax:add} (efficiency and additivity), Shapley values are a model averaging procedure, being the weighted average~\eqref{eq:shapley} of marginal contributions. In statistics, model averaging has been used to combine the strengths of several candidate selected models~\cite{[G10]}. These candidate models may arise from perturbations, e.g., when the result of a model selection procedure is recognised to be sensitive to sample effects or other conditions, or in circumstances where there is no clear optimum of the evaluation function. In any case, model averaging procedures are traditionally not used for feature selection, but to give a weighted average of estimates or predictions associated with a number of different models. 

There is a compelling reason for caution around the direct use of model averaging for feature selection: The average performance of a feature across \textit{all} submodels may not be indicative of the particular performance of that feature in the set of optimal submodels. Ideally, one would select all features explicitly on the weight of their contribution to submodels that are optimal. We illustrate this with the following example.

\begin{example}[\textit{Taxicab payoff}] \label{ex:taxicab}
Consider a game with $\dimension = 3$ players and ``taxicab'' style payoff with characteristic function given by
\begin{equation}
   \CF(S) = \left\{ 
   \begin{array}{ll}
   10, & \text{if $3 \in \submodel$}, \\
   7, & \text{if $3 \not\in \submodel$ and $2 \in \submodel$}, \\
   3, & \text{if $2,3 \not\in \submodel$ and $1 \in \submodel$}, \\
   0, & \text{otherwise.}
   \end{array}
   \right.
\end{equation}
This game can be pictured as one taxicab ride, where the homes of players~$1$ and~$2$ lie on route to the home of player~$3$. From the driver's perspective, the maximum profit is obtained from player~$3$, regardless of any absence of players~$1$ or~$2$. If this is a feature selection task, where all features have equal cost, then the optimal model does not contain features $1$ or $2$. From a model selection point of view, players $1$ and $2$ are useless features. However, since the Shapley values are
\[
\varphi = (\varphi_1, \varphi_2, \varphi_3) = (1,3,6) \,,
\]
from a \textit{fairness} point of view, in the sense of the Shapley value, the players $1$ and $2$ are not worthless, since they add value to at least one other set of players (at least to the empty set, in the case of player $1$).
\end{example}

In~\Cref{ex:taxicab}, players $1$ and $2$ are not ``null players", in the sense of the~\Cref{ax:nul}, but they contribute nothing to the optimal model. As the example demonstrates, performance of features may increase within submodels due to an interaction with a dominant feature. Averaging across all submodels takes these superfluous performances into account, when ideally the presence of the dominant feature should be fixed. Indeed, a possible solution to this problem may be to identify and fix the presence of such dominating features, prior to computing the Shapley value -- though at this stage feature selection may no longer be required. A concrete manifestation of~\Cref{ex:taxicab} is explored in~\Cref{sim:taxicab}.

\subsection{The meaning of efficiency} \label{sec:mean-eff}

\Cref{ax:eff} (efficiency), states that the evaluation of the full model, $C(\fullmodel)$, should be distributed losslessly amongst the features. This axiom narrows the scope of possible model averaging procedures to those that treat the full model, not the selected model, as the final outcome. When the objective function is non-monotonic, this becomes especially distinguished from the problems discussed in~\Cref{sec:mean-avg}. Since the full model is not generally the target, non-monotonic evaluation functions imply that \textit{efficient payoffs may waste value}.

Non-monotonic evaluation functions (\Cref{def:monotonic}) are those that may decrease as the number of features increases, i.e., there exist submodels $\submodeltwo, \submodel$ such that $\CF(T) < \CF(S)$ for some $T \supset S.$ This means, for example, that we may have
\[
\sum_{i \in \fullmodel} \shapley_i = \CF(\fullmodel) < \CF(\submodel),
\]
for some $\submodel \subset \fullmodel.$ In this case, the Shapley values do not sum to the maximum possible value of the feature set, over all subsets of features. In other words, the Shapley values sum to the payoff for the full model, but they don't sum to a payoff that is optimal. Note that, at least for the exact Shapley value, evaluation of all $2^\fullmodel$ submodels is an intermediate step in the calculation. From these evaluations, the optimal model can be computed. In practice, for some approximation to \textit{optimal efficiency}, it may be preferable to estimate the optimal model prior to approximating the Shapley value -- at which stage feature selection is no longer required.

\subsection{The meaning of balanced contributions} \label{sec:mean-bal}

\Cref{ax:bal} (balanced contributions), can be substituted for Axioms \ref{ax:nul}--\ref{ax:add} (null player, symmetry and additivity). \Cref{ax:bal} captures a notion of fairness in the rewards of cooperation between players in a \ac{TU} game. The symmetry axiom alone may be undesirable in certain contexts. In the case of two strongly correlated features, the symmetry axiom dictates that both should receive approximately the same attribution. However, in feature selection, the high correlation implies that one of the two features is redundant. Here, we regard a feature $X_\pone$ as \textit{redundant} in the presence other features $\mathcal{X} = \set{X_\ptwo,\, \ptwo \in \submodel}$, if feature $X_\pone$ is conditionally independent of $Y$, conditional on $\mathcal{X}$, for which we write $Y \ind X_\pone | \mathcal{X}.$ Redundancy is investigated in more precise contexts in \Cref{sim:markov1,sim:markov2}.

A second consequence that may be attributed to \Cref{ax:bal}, is that the earnings received by a coalition, after discounting the earnings of its subcoalitions, are shared equally amongst the players in that coalition. To better understand this, see the formulation of the Shapley value in terms of Harsanyi dividends~\cite{[G1]}, which we do not enter into here. In particular, from \eqref{eq:shapley}, a player's contribution to teams of size $k$ is averaged across all $\binom{\dimension}{k}$ teams of size $k$. Since the binomial coefficient decreases in $|k - \dimension/2|$ for fixed $\dimension$, a player's single contribution to a team at the extremes of the spectrum of team sizes will be weighted higher than to team sizes close to $\dimension/2$. As a consequence, poor performance of a feature on particularly small submodels, such as the singleton models, may be weighted highly even if such small submodels are not attractive for the model selection task. We illustrate this in~\Cref{ex:secret}

\begin{example} \label{ex:secret} 
One should be wary of deciding that the features with high Shapley values are also the strongest contributors to model performance. Consider the scenario with $\dimension = 3$ and a ``secret holder'' style payoff where player $1$ is alone worthless, but has the ``secret'' that endows any team with the maximum possible payoff. Specifically, suppose we have the payoff lattice in~\Cref{fig:lattice}.

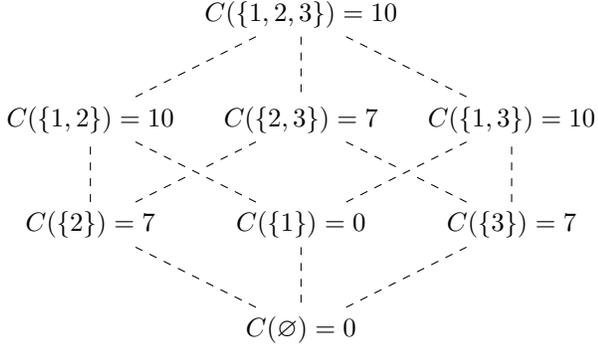
\begin{figure}
\begin{center}
\begin{tikzpicture}[scale=0.7]
  \node (max) at (0,4) {$\CF(\{1,2,3\}) = 10$};
  \node (a) at (-4,2) {$\CF(\{1,2\}) = 10$};
  \node (b) at (0,2) {$\CF(\{2,3\}) = 7$};
  \node (c) at (4,2) {$\CF(\{1,3\}) = 10$};
  \node (d) at (-4,0) {$\CF(\{2\}) = 7$};
  \node (e) at (0,0) {$\CF(\{1\}) = 0$};
  \node (f) at (4,0) {$\CF(\{3\}) = 7$};
  \node (min) at (0,-2) {$\CF(\varnothing) = 0$};
  \draw[dashed] (min) -- (d) -- (a) -- (max) -- (b) -- (f)
  (e) -- (min) -- (f) -- (c) -- (max)
  (d) -- (b);
  \draw[dashed, preaction={draw=white, -,line width=6pt}] (a) -- (e) -- (c);
\end{tikzpicture}
\end{center}
\caption{A diagram of the evaluation function with ``secret holder'' style payoff described in~\Cref{ex:secret}.}
\label{fig:lattice}
\end{figure}

The Shapley value vector is
\[
\varphi = (\varphi_1, \varphi_2, \varphi_3) = (2,4,4).
\]
The players $2$ and $3$ are attributed twice the value of player $1$, but if the submodel $\{2,3\}$ is selected, then performance will be suboptimal. Furthermore, from the full model $\{1,2,3\}$, we would do well to discard player $2$ or $3$. This example is investigated further in \Cref{sim:secret}.
\end{example}

\section{Experimentation} \label{sec:experiments}

In this section, where the \ac{DGP} is known to us, we investigate, through simulation, a number of situations in which the results produced by~\Cref{alg:attsel} may be undesirable. In~\Cref{sim:markov1,sim:markov2}, we reproduce the results of~\cite[Theorem 8 and 9]{[F1]} and extend them via simulation to include the \texttt{SHAP~FSelection} formulation~\cite{[F14]} (i.e., feature selection by ranking mean absolute \ac{SHAP} values of model predictions), and the \ac{SAGE} formulation. Accurate definitions of \ac{SHAP} and \ac{SAGE} methods are very detailed, so rather than reproducing them here, we encourage the reader to access~\cite{[S5]} (\ac{SHAP}) and~\cite{[F15]} (\ac{SAGE}). 

\subsection{Markov boundary experiment 1} \label{sim:markov1}

\begin{figure}
    \begin{subfigure}[b]{0.49\textwidth}
        \includegraphics[width=\textwidth]{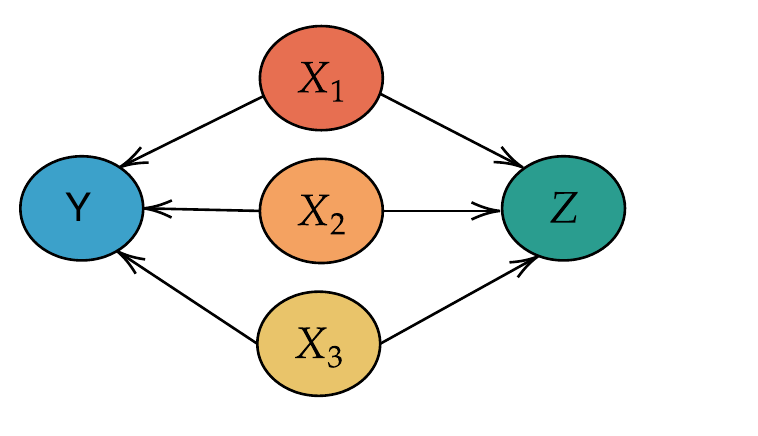}
    \caption{\label{fig:ma1}}
    \end{subfigure}
    \begin{subfigure}[b]{0.49\textwidth}
        \centering
        \includegraphics[width=0.6\textwidth]{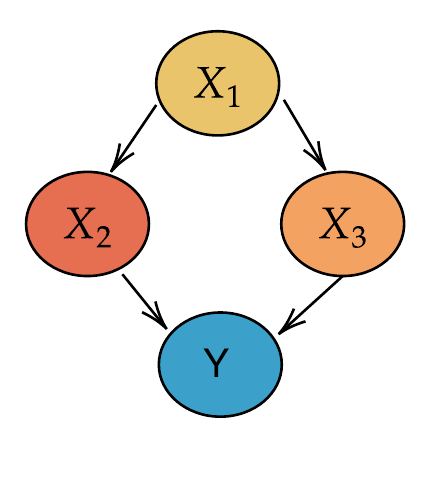}
        \caption{\label{fig:ma2}}
    \end{subfigure}
    \caption{Causal graphs, where (\protect\subref{fig:ma1}) corresponds to the \ac{DGP}~\eqref{eq:ma8dgp} and (\protect\subref{fig:ma2}) corresponds to~\eqref{eq:ma9dgp}.}
\end{figure}

In this experiment we consider a scenario where we wish to predict a response variable $Y$ from four features $X_1, X_2, X_3, Z$, with the following \ac{DGP} suggested in \cite[Theorem 8]{[F1]},
\begin{equation}  \label{eq:ma8dgp}
    \begin{split}
        \,X_1,\;& X_2,\; X_3 \sim \mathcal{N}(0,4), \\
        Y =& \,X_1 + X_2 + X_3 + \eps, \\
        Z =& \,X_1 + X_2 + X_3 + \gamma, \\
    \end{split}
\end{equation}
where $\eps, \gamma \sim \mathcal{N}(0, 4)$ introduce irreducible uncertainty into the relationships defining $Y$ and $Z$, respectively, via normal perturbations with means zero and variances $4$. The causal graph is shown in~\cref{fig:ma1}, where  $X_1, X_2, X_3 \to Y$ and $X_1, X_2, X_3 \to Z$. Regardless of any implied causal relationships, we can interpret it as a case where $Z$ is separated from $Y$ via two terms of uncertainty ($\eps$ and $\gamma$), while the set $\{X_1, X_2, X_3\}$ is separated from $Y$ by only one term of uncertainty ($\eps$). It follows that $Y$ is conditionally independent of $Z$, given the values $\{X_1, X_2, X_3\}$, but not vice versa. Furthermore, $\{X_1, X_2, X_3\}$ is the minimal set with the property that the remaining features are conditionally independent of $Y$. Thus, $X_1, X_2, X_3$ are referred to as \textit{Markov boundary members} of the \ac{DGP}. However, as also shown by~\cite{[F1]}, when we employ the $R^2$ evaluation function, the Shapley values are
\[
    (\phi_Z, \phi_{X_1}, \phi_{X_2}, \phi_{X_3}) = (0.26, 0.16, 0.16, 0.16) \,,
\]

The three features $X_1, X_2, X_3$, the Markov boundary members, all have smaller Shapley values than $Z$, the non-Markov boundary member. This is not a peculiarity of that particular game formulation. Simulating a data set with sample size $n=10^6$ from \ac{DGP} \eqref{eq:ma8dgp}, and training an \texttt{XGBoost} regression model, the \texttt{SHAP FSelection} method sorts the features as $(Z, X_1, X_2, X_3)$ with corresponding mean absolute \ac{SHAP} values $(1.4, 1.1, 1.1, 1.1)$. In both cases, a top-$3$ feature selection procedure will select $(Z, X_1, X_2)$, rather than the preferred triple $(X_1, X_2, X_3)$, thus introducing an unnecessary term of uncertainty ($\gamma$) into the model.

The \ac{SAGE} values, on the other hand, which compute a global Shapley value of model loss, representing the predictive power associated with each feature in a model, produce feature importance scores of $(1,4,4,4)$ for the features $(Z, X_1, X_2, X_3)$. Thus, while the \texttt{SHAP FSelection} and $R^2$ formulations produce poor results for model selection, the \ac{SAGE} values successfully highlight the appropriate features in this scenario.

\subsection{Markov boundary experiment 2} \label{sim:markov2}

We study a generalisation of a \ac{DGP} suggested by \cite[Theorem 9]{[F1]}, who consider the special case $\ell=0.05$. First, we sample feature $X_1$ uniformly from $\set{1,2,3,4}$. Then, $X_1,X_2,X_3$ are sampled as follows.
\begin{equation}\label{eq:ma9dgp}
    \begin{split}
        P (X_2 = 1|X_1 = 1) = \ell,& \;   P (X_2 = 1|X_1 = 3) = \ell-1,     \\
        P (X_2 = 1|X_1 = 2) = \ell,& \;   P (X_3 = 1|X_1 = 2) = \ell-1,    \\
        P (X_3 = 1|X_1 = 1) = \ell,& \:   P (X_2 = 1|X_1 = 4) = \ell-1,  \\
        P (X_3 = 1|X_1 = 3) = \ell,& \;   P (X_3 = 1|X_1 = 4) = \ell-1,  \\
        P (Y = 1|X_2 &= 0, X_3 = 0) = 0.9,   \\
        P (Y = 1|X_2 &= 0, X_3 = 1) = 0.05,  \\
        P (Y = 1|X_2 &= 1, X_3 = 0) = 0.15,  \\
        P (Y = 1|X_2 &= 1, X_3 = 1) = 0.9 \,. \\
   \end{split}
\end{equation}
Here, $\ell \in (0,1)$. The causal graph is depicted in~\cref{fig:ma2}. In~\cite{[F1]}, the evaluation function $m$ is used, defined as, 
\begin{equation}
m(S) = \sum_{\mathbf{x}_S}P(\mathbf{X}_S=\mathbf{x}_S) V(S),
\end{equation}
\[
V(S) = \max\set{P(Y=1|\mathbf{X}_S = \mathbf{x}_s),P(Y=0|\mathbf{X}_S = \mathbf{x}_s)},
\]
where $\mathbf{X}_\submodel$ is the vector of features indexed by $\submodel$. The resulting Shapley values, as given in \cite{[F1]}, with $\ell=0.05$ are
\[
        (\phi_{X_1}, \phi_{X_2}, \phi_{X_3}) = (0.22, 0.09, 0.09) \,.
\]

Here, despite introducing unnecessary irreducible error into the model (see \Cref{fig:ma2}), the non-Markov boundary variable $X_1$ is given the highest preference. Simulating a data set of sample size $n=10^6$, and training a simple \texttt{XGBoost} classification model, for $\ell = 0.05$, the \texttt{SHAP FSelection} method also sorts the features as $(X_1, X_2, X_3)$, with respective mean absolute \ac{SHAP} values $(1.43, 0.50, 0.40)$. On the other hand, \ac{SAGE} sorts the features as $(X_2,X_3,X_1)$ with values $(0.0797, 0.0787, 0.0003)$. Thus, as in~\Cref{sim:markov1}, the \ac{SAGE} values sidestep the issues of the \texttt{SHAP FSelection} and $m$ formulations, instead producing a correct ranking for feature selection.

\begin{figure}
      \begin{subfigure}[b]{0.9\textwidth}
        \centering
        \includegraphics[width=\textwidth]{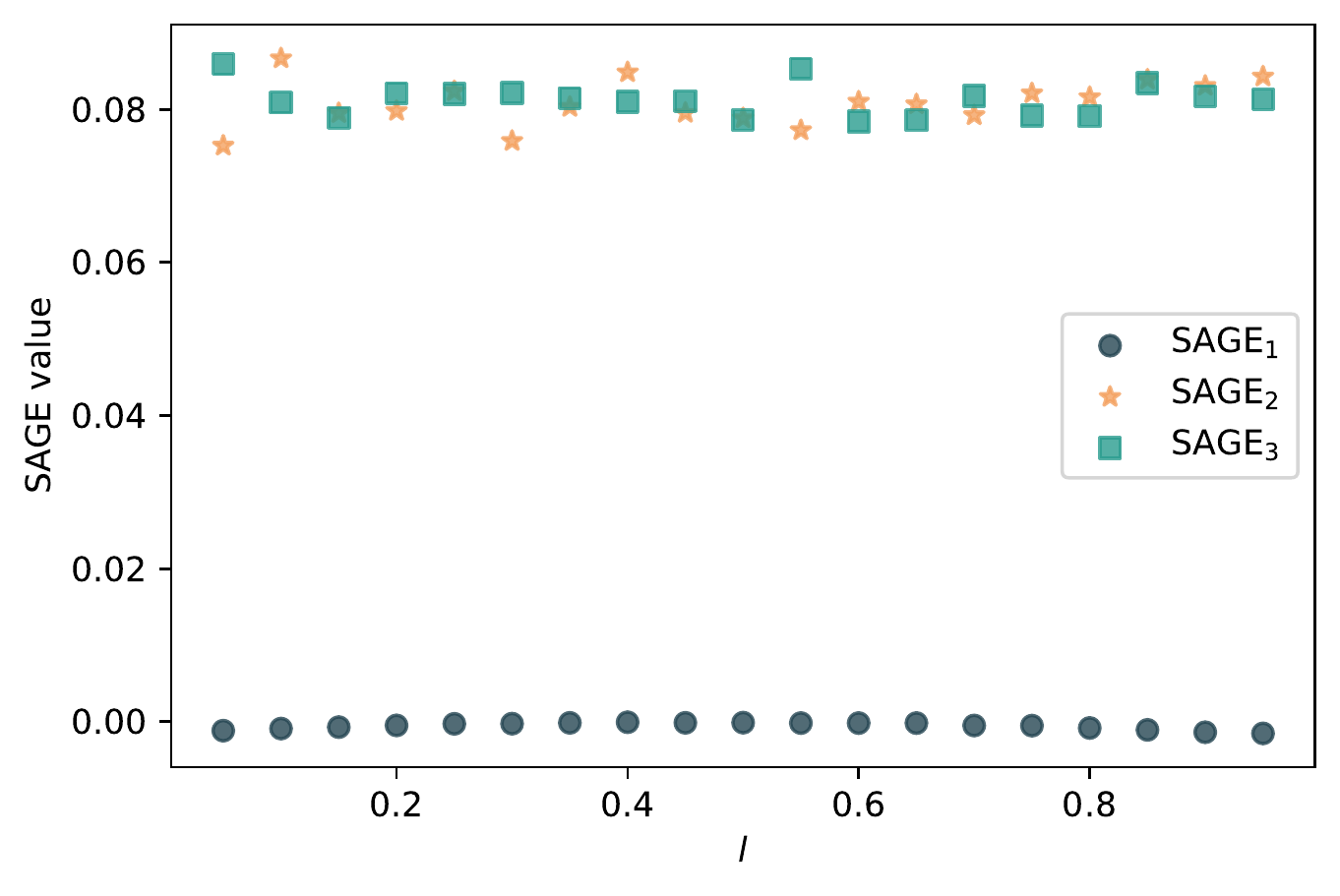}
        \caption{\label{fig:sage_th9}}
    \end{subfigure}
    \begin{subfigure}[b]{0.9\textwidth}
        \centering
        \includegraphics[width=\textwidth]{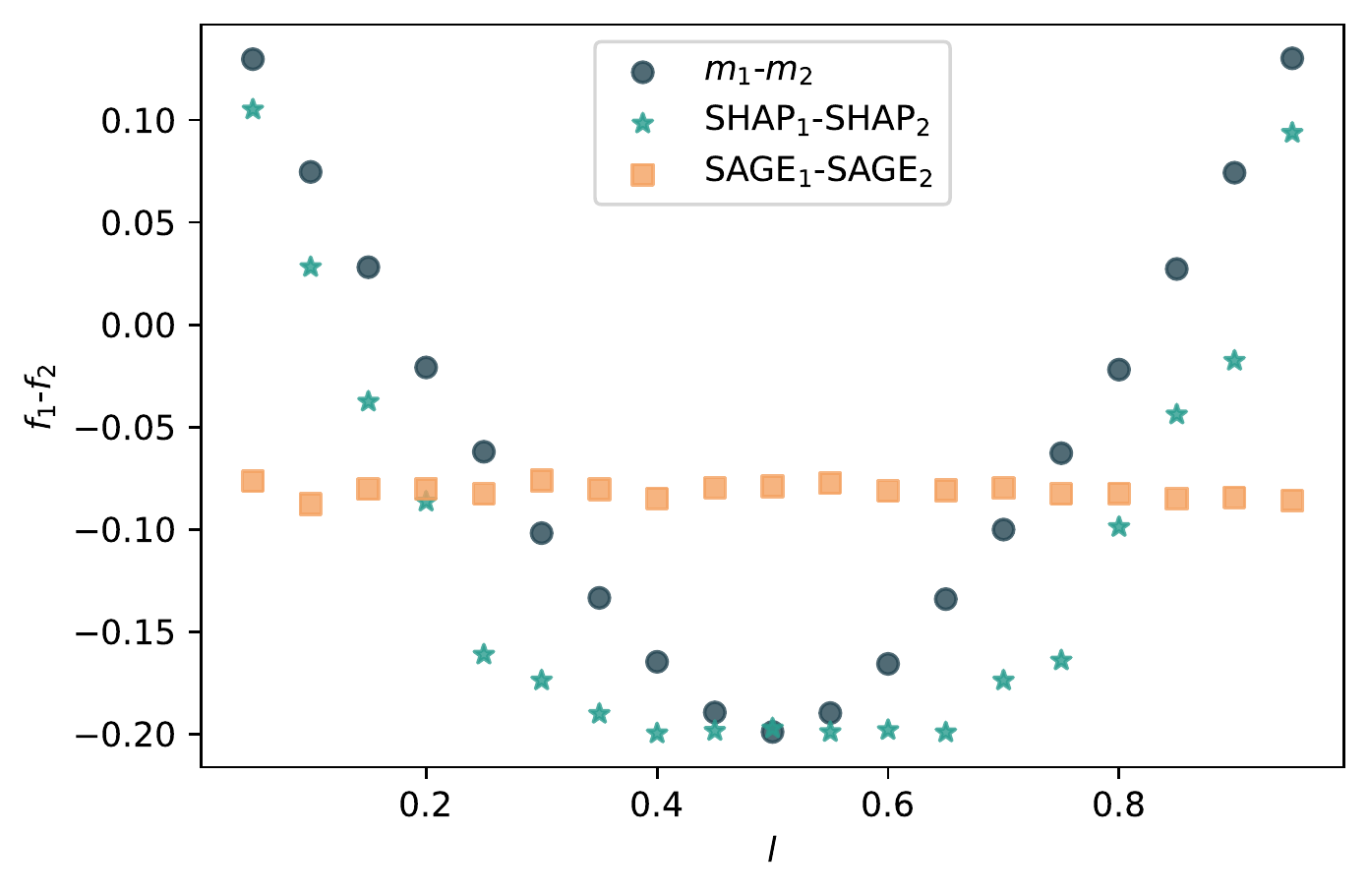}
        \caption{\label{fig:sage_vs_shapley_th9}}
    \end{subfigure}
    \caption{The (\protect\subref{fig:sage_th9}) \ac{SAGE} values for the \ac{DGP} in~\eqref{eq:ma9dgp} across a grid of $0.05~\leq~\ell~\leq~0.95$, (\protect\subref{fig:sage_vs_shapley_th9}) Shapley values for the mean absolute \ac{SHAP}, \ac{SAGE} and $m$ function formulations, for the difference $\shapley_1 - \shapley_2$ between attribution to $X_1$ and $X_2$, across the same range of $\ell$ values. Similar results were observed for the difference $\shapley_1 - \shapley_3.$ Values $\shapley_1 - \shapley_2 < 0$ indicate regions of the parameter space for which model selection results are inadequate.} 
\end{figure}

To determine the relationship of the parameter $\ell \in (0, 1)$ to the pathology, we simulate $20$ more data sets, equally spaced on the grid $0.05 \leq \ell \leq 0.95$, and calculate \texttt{SHAP FSelection} and \ac{SAGE} values for each value of $\ell$. The variation of the \ac{SAGE} values with $\ell$ is shown in~\cref{fig:sage_th9}. Calculating the differences $\varphi_1 - \varphi_2$ in Shapley values of the variables $X_1$ and $X_2$, for the \ac{SHAP}, \ac{SAGE} and $m$ formulations, yields~\cref{fig:sage_vs_shapley_th9}. Similar behaviour is realised in the differences $\varphi_1 - \varphi_3$. From this, we see that \ac{SAGE} performs admirably over the investigated parameter space, while the \ac{SHAP} and $m$ formulations perform poorly for approximately $|\ell-1/2| > 0.3$.

\subsection{A secret holder experiment} \label{sim:secret}

In this experiment we consider the \ac{DGP}
\begin{equation} \label{dgp:secret}
Y = \sum_{i=1}^3 \beta_i X_i + \sum_{i=1}^3\sum_{j=1}^3 \beta_{ij}X_iX_j + \varepsilon,
\end{equation}
with  $\varepsilon \sim \mathcal{N}(0,1)$ and, given parameters $t_1,t_2 \in \R,$ having $\beta_1 = 0, \beta_2 = \beta_3 = t_1 \neq 0, \beta_{23} = 0, \beta_{12} = \beta_{13} = t_2 \neq 0$. The features $X_i, i \in \{1,2,3\}$ are generated as independent standard normal random variables.

\begin{figure}
    \centering
    \includegraphics[width=0.9\textwidth]{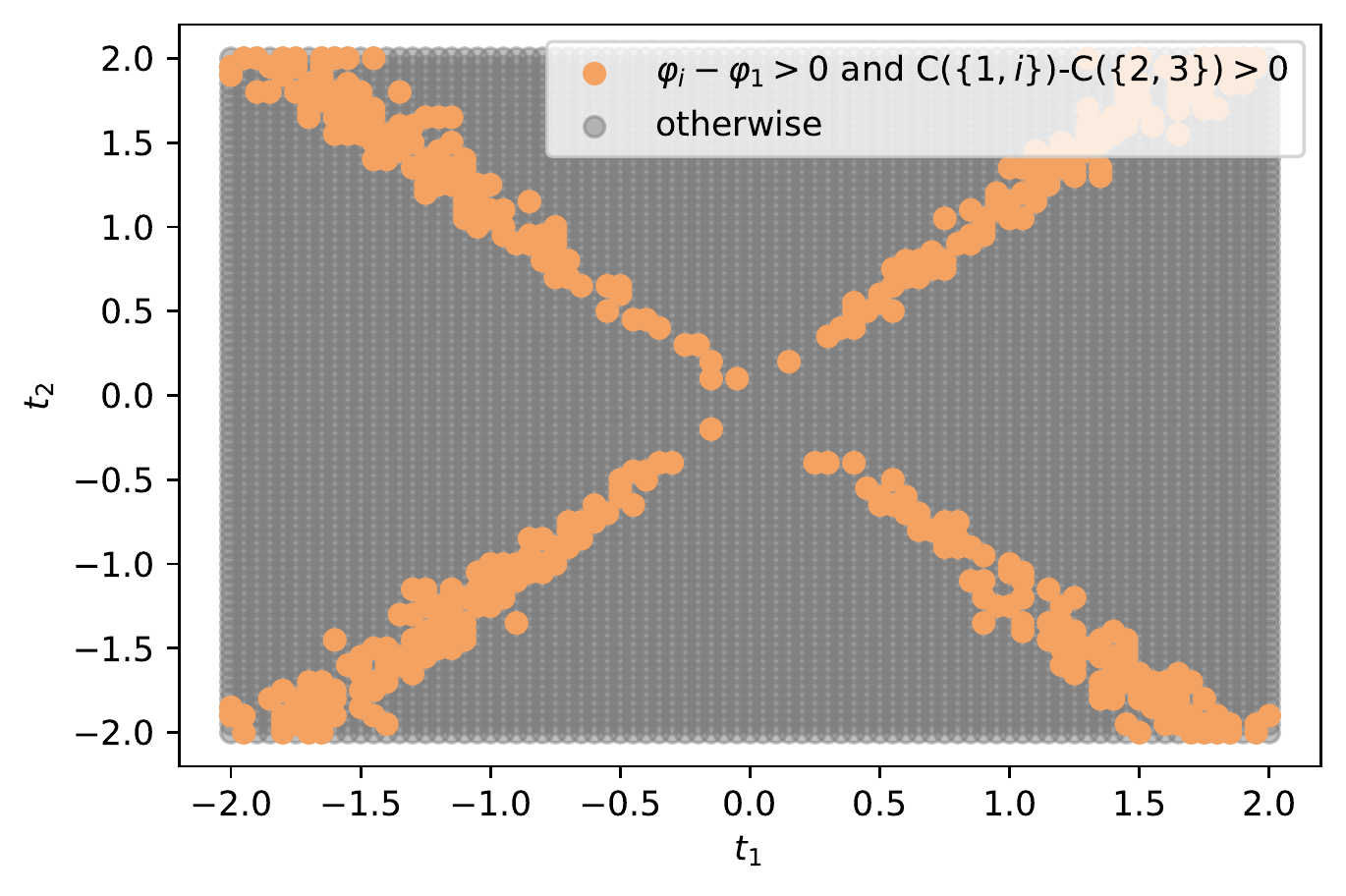}
    \caption{\label{fig:t1_vs_t2_diffs} 
    Grid points $(t_1,t_2)$ for $-2 \leq t_i \leq 2$ at $0.05$ increments, in the parameter space for the \ac{DGP} \eqref{dgp:secret} described in \Cref{sim:secret}. Points in orange are where $\shapley_i - \shapley_1 > 0$, as well as $\CF(\{1,i\}) - \CF(\{2,3\}) > 0$, for both $i = 2$ and $i = 3$. 
    }
\end{figure}

From \eqref{dgp:secret}, we simulate a total of $6561$ data sets of sample size $n=1000$, producing one data set for each position on the $81\times81$ parameter grid $(t_1,t_2)$ with $-2 \leq t_i \leq 2$ and increments of length $0.05$. On each data set we compute Shapley values for the conditional log likelihood evaluation function 
\begin{equation}\label{cf:lik}
L(S) = \mathcal{L}(\theta; \mathbf{x}_\varnothing) - \mathcal{L}(\theta; \mathbf{x}_S) \,,
\end{equation}
where $\theta$ is estimated via least squares, using the closest submodel of the true model \eqref{dgp:secret}, given the vector $\mathbf{x}_S$ of features indexed by $S$. In \Cref{fig:t1_vs_t2_diffs} we present the results of highlighting all $(t_1,t_2)$ such that the characteristic function is pathological in the sense of matching \Cref{ex:secret}. From this we see that pathologies occur at approximately $t_2 = \pm(|t_1|+\alpha)$ for $0 < \alpha < 0.4.$

Choosing $t_1 = 2, t_2 = 2.2$, we compute the characteristic function for \eqref{cf:lik}:
\[
\begin{split}
\CF(\{1,2,3\}) & = 1.52, \\
\CF(\{1,2\}) = 0.34, & \; \CF(\{1\}) = 0.00, \\
\CF(\{2,3\}) = 0.27, & \; \CF(\{2\}) = 0.13, \\
\CF(\{1,3\}) = 0.34, & \; \CF(\{3\}) = 0.11, \\
\end{split}
\]
which we compare to \Cref{fig:lattice}. The Shapley values are
\[
(\shapley_1, \shapley_2, \shapley_3) = (0.5,0.52,0.51).
\]
Thus, although the results are quite close, we see that $X_2$ and $X_3$ are favoured in feature selection, despite $X_1$ being the ``secret holder'' and a member of both optimal submodels of size $2$. 

The corresponding mean absolute \ac{SHAP} values are $(0.97, 1.52, 1.48)$ and the \ac{SAGE} values are $(4.6, 5.8, 5.9)$. \ac{SHAP} and \ac{SAGE} disagree regarding whether feature $X_2$ or $X_3$ is the most important, but neither method gives precedence to $X_1$, which is in fact most important.

\subsection{A taxicab experiment} \label{sim:taxicab}
The following scenario is a concrete realisation of \Cref{ex:taxicab}. Consider $\dimension$ variables generated as
\begin{equation}
    X_i = \mathcal{N}(0,1) + a_i \,, 
\end{equation}
where $a_1 < a_2 < \dots < a_\dimension$. The generative model is
\begin{equation}
    Y = \text{max}\{X_1, X_2, \dots, X_\dimension \} + \eps \,,
\end{equation}
where $\eps \sim \mathcal{N}(0,1)$. The predictive model that we apply to this data is the correct model, which is simply
\[
    \hat{Y} = \text{max}\{X_1, X_2, \dots, X_\dimension \}\,.
\]
We evaluate this model using as evaluation function the following difference between the mean squared errors
\[
    \text{MSE}(y, 0) - \text{MSE}(y,\hat{y}) \,,
\]
and choosing $\dimension=3$ with $(a_1,a_2, a_3) = (5,10,20)$. The characteristic function evaluates to
\[
\begin{split}
\CF(\{1,2,3\}) & = 546, \\
\CF(\{1,2\}) = 371, & \; \CF(\{1\}) = 210, \\
\CF(\{2,3\}) = 546, & \; \CF(\{2\}) = 371, \\
\CF(\{1,3\}) = 546, & \; \CF(\{3\}) = 546. \\
\end{split}
\]
The resulting Shapley values are 
\[
    (\varphi_1, \varphi_2, \varphi_3) = (70, 151, 325) \,.
\]
We see that the scenario in \Cref{ex:taxicab} has been generated in this more concrete setting.
\section{Discussion}

Our investigations of the axioms in~\Cref{sec:mean} prompted a number of experiments on potentially pathological \ac{DGP}s, presented in~\Cref{sec:experiments}. In each experiment, we were able to define a reasonable evaluation function and game formulation for which the Shapley value behaved in an undesirable way for feature selection. When these experiments were applied to the mean absolute \ac{SHAP} (i.e., \texttt{SHAP FSelect}) and \ac{SAGE} formulations, the former performed poorly in all of the three experiments (\Cref{sim:markov1,sim:markov2,sim:secret}), while the latter performed favourably in two out of those three experiments.

Our results confirm that the axioms do not \textit{in general} provide any guarantee that the Shapley value is suited to feature selection, and may in some cases imply the opposite. However, more importantly, the relevance of the Shapley value to the feature selection task (indeed, to any \ac{ML} task) is governed by the specific game formulation, and the justification of this relevance from the axioms is non-trivial. Crucially, it is the authors' perception that abstract general axioms presented as ``favourable and fair" may introduce a significant potential for \textit{magical thinking} \cite{[O4]} in \ac{XAI}. It is our intention to caution against this, and to emphasize the nuance in any application of Shapley values.

There are a large variety of alternatives to the Shapley value provided in the literature on game theory \cite{[G1]}. As an example, \cite[p.55]{[G1]} suggests (under Games with Hierarchies), an \textit{inessential player} axiom, dictating that players that are not essential should receive zero value.

Future work is needed to thoroughly explore the application of game theoretic solution concepts to feature selection and attribution. Extensive empirical studies are needed to understand the game formulations and axioms that are suited to the large variety of practical and frequently occurring feature selection tasks in \ac{XAI} and \ac{ML}.

\bibliography{bibliography}
\bibliographystyle{IEEEtran}

\end{document}